\documentclass{article}
\usepackage{spconf,amsmath,graphicx}
\usepackage{soul}
\usepackage{amssymb}
\usepackage[usenames]{color}
\usepackage{makecell}
\usepackage{hyperref}
\usepackage{multirow}
\usepackage[table]{xcolor}
\usepackage{enumitem}
\usepackage{rotating}
\usepackage{array}
\usepackage{caption}
\usepackage{subcaption}
\usepackage{booktabs}
\definecolor{lavender}{rgb}{0.9, 0.9, 0.98}
\definecolor{platinum}{rgb}{0.9, 0.89, 0.89}

\usepackage{bbding}
\title{Facial Kinship Verification from remote photoplethysmography}

\name{Xiaoting~Wu$^{\star}$, Xiaoyi Feng$^{\dagger}$, Constantino Álvarez Casado $^{\star}$, Lili Liu$^{\dagger \star}$, and  Miguel Bordallo López$^{\star}$\textsuperscript{\Envelope}  \thanks{This research was supported by the Research Council of Finland (former Academy of Finland) 6G Flagship Programme (Grant Number: 346208). CSC-IT Center for Science, Finland, is acknowledged for providing computational resources. Bordallo López Miguel is the corresponding author: miguel.bordallo@oulu.fi.}}
\address{$^{\star}$ Center for Machine Vision and Signal Analysis, University of Oulu, Oulu, Finland \\
      $^{\dagger}$  School of Electronics and Information, Northwestern Polytechnical University, Xi’an, China}
\begin{document}
\topmargin=0mm
%\ninept
\maketitle
\begin{abstract}
Facial Kinship Verification (FKV) aims at automatically determining whether two subjects have a kinship relation based on human faces. It has potential applications in finding missing children and social media analysis. Traditional FKV faces challenges as it is vulnerable to spoof attacks and raises privacy issues. In this paper, we explore for the first time the FKV with vital bio-signals, focusing on remote Photoplethysmography (rPPG). rPPG signals are extracted from facial videos, resulting in a one-dimensional signal that measures the changes in visible light reflection emitted to and detected from the skin caused by the heartbeat. 
Specifically, in this paper, we employed a straightforward one-dimensional Convolutional Neural Network (1DCNN) with a 1DCNN-Attention module and kinship contrastive loss to learn the kin similarity from rPPGs. 
The network takes multiple rPPG signals extracted from various facial Regions of Interest (ROIs) as inputs. Additionally, the 1DCNN attention module is designed to learn and capture the discriminative kin features from feature embeddings. 
Finally, we demonstrate the feasibility of rPPG to detect kinship with the experiment evaluation on the UvANEMO Smile Database from different kin relations.
\end{abstract}
%
%%%%%%%%%%%%%%%%%%%%%%%%%%%%%%%%%%%%%%%%%%%%%%%%%%%
\vspace{0mm}
\section{Introduction}
\vspace{0mm}
\label{sec:intro}
%Kinship verification 
%motivation for rPPG
%motivation for proposed method
%related work on kinship
%contribution
Biometrics provides a way of recognizing human kinship using biometric traits, such as facial images~\cite{dibeklioglu2017visual}, voice~\cite{wu2022audio} and gait~\cite{bekhouche2020kinship}, which is known as Facial Kinship Verification (FKV)~\cite{wu2022facial}. It has various potential applications in genetic and anthropological studies, social security, and locating missing children. However, with the growing of generative models, the threats of malicious attacks on biometric traits exhibit potential risks for the trustworthy system~\cite{cai2021generative}. For example, voice can be imitated, and generated kin facial images~\cite{li2023stylegene} can be used as spoof inputs to the system. Thus, it is essential to explore the robust and new biometric traits for the FKV problem.

\begin{figure}[t!]
\centering
\includegraphics[width=0.45\textwidth]{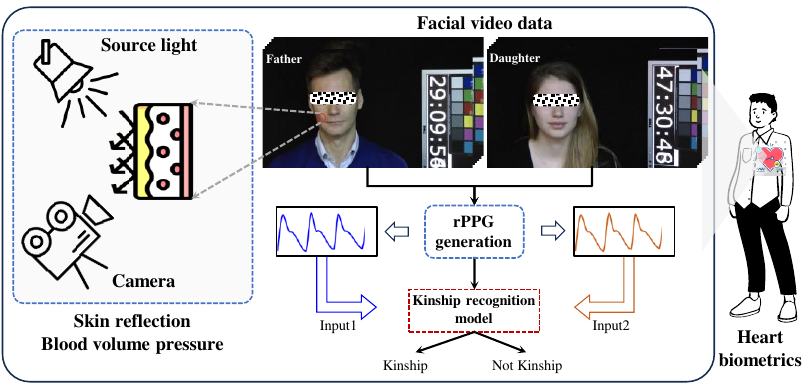}
\caption{Facial Kinship Verification with Remote Photoplethysmography. We first measure the reflected color changes from facial skin areas resulting in the rPPG signals. Then, the rPPG that includes heartbeat biometric is used as the input to the kinship verification system.}
\label{fig:motivation}
\vspace{0mm}
\end{figure}

The blood pumped by the heartbeat results in electrical cardiac activities in the skin that can be trustworthy inputs for robust biometric systems~\cite{patil2018non}, measured by electrocardiogram (ECG)~\cite{huang2016robust} and electroencephalogram (EEG)~\cite{poulos1999person}. On the other hand, heartbeats cause capillary dilation and constriction, which effects as detectable changes in the skin's reflection of visible light~\cite{boccignone2022pyvhr}. Photoplethysmography (PPG) is a technique that quantifies these changes in reflected light from the skin, thereby capturing the dynamics of heartbeats and providing insights into cardiovascular parameters, such as blood pressure variantions~\cite{patil2018non}.

Studies as early as 1925~\cite{heart1925} demonstrated that heartbeats convey genetic information. Since then, many studies~\cite{russell1998heritability,singh2012bioelectrical} have shown that Electrocardiogram (ECG) analysis of twins presents more significant intra-pair similarities than those of unrelated individuals. In recent years, physiological signals such as ECG, Photoplethysmography (PPG), remote PPG, and Phonocardiogram (PCG) have been widely used in automatic biometric recognition, such as HeartID~\cite{fatemian2010heartid} and Heart Biometric~\cite{rathore2020survey}. However, the study of FKV using physiological signals remains largely unexplored.

\begin{figure*}[t!]
\centering
\includegraphics[width=0.95\textwidth]{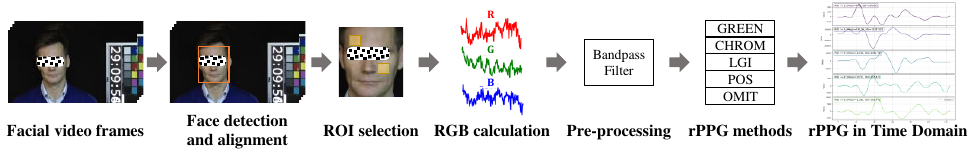}
\caption{The general pipeline of the rPPG measurement~\cite{boccignone2022pyvhr,face2ppg}. The input is a sequence of facial video frames, with the output signals of rPPG data that implicitly reflect the blood volume changes reflected by the facial skin. Several steps are implemented to generate the rPPG: 1) face detection and alignment, normalizing the cropped faces; 2) ROI selection, choosing the skin regions that contain the PPG-related information; 3) RGB calculation, extracting the raw signal from color spaces; 4) pre-processing, to filter out the noisy data with the frequency filter or standard normalization. 5) rPPG methods, recovering the skin color variations into physiological signals.}
\label{fig:ppg}
\vspace{0mm}
\end{figure*}

Remote Photoplethysmography (rPPG) estimates the blood flow from facial videos by detecting subtle changes in skin color associated with the cardiac cycle. This non-invasive and non-contact technique, which captures unique physiological patterns, has sparked curiosity about its potential applications beyond mere heart rate estimation. Inspired by biometric studies, in this paper, we pose the question: \textit{Is it possible to verify kinship using remote Photoplethysmography (rPPG)}? To answer this question, we present a proposed pipeline as shown in Fig.~\ref{fig:motivation}. 

As the pioneering study in this area, we carefully designed our experiment based on the clear, front-facing UvANEMO Smile Database~\cite{dibekliouglu2012you} that was collected under controlled indoor conditions. First, five rPPG methods are employed to measure the rPPG signals from facial videos with an unsupervised-learning manner. To perform a comparison between rPPG signals from the subject pair, we proposed a 1DCNN-Attention model. 1DCNN is a typical deep learning method for processing one-dimensional sequence data~\cite{wang2017time}. To enhance the robustness of the network and mitigate the variance of different ROIs, we use multiple rPPG signals as inputs for each subject, resulting in a multi-channel data configuration at the input layer. Furthermore, the 1DCNN channel attention module is applied to learn discriminative kin features. The network employs a Siamese-like architecture for the training. Finally, the contrastive loss is added to guide model training.

Our contributions can be summarized as follows:
\begin{itemize}[itemsep=2pt,topsep=0pt,parsep=0pt]
\item We explore the remote photoplethysmography for kinship recognition and formulate the problem for the first time.
\vspace{2mm}
\item We introduce a 1DCNN-Attention model to adaptively capture kin-related features from rPPGs for kinship verification.
\vspace{2mm}
\item Through experiments on the UvANEMO Smile Database \cite{dibekliouglu2012you}, we demonstrate the feasibility of verifying kinship using rPPGs and establish the first benchmark for this study.
\end{itemize}

\vspace{0mm}
%%%%%%%%%%%%%%%%%%%%%%%%%%%%%%%%%%%%%%%%%%%%%%%%%%%
\section{Related works}% on Facial Kinship Verification
\vspace{0mm}
%\textbf{Facial Kinship Verification}
The first study on kinship verification within the computer vision field was carried out in 2010~\cite{fang2010towards}, utilizing a variety of facial geometric and color attributes to determine appearance similarities between two subjects. After 2010, efforts largely concentrated on employing identity features extracted from images and videos for the FKV problem~\cite{wu2022facial}. Traditional methods, such as hand-crafted features~\cite{lu2014neighborhood}, have the advantage of computational efficiency, especially in small datasets. However, traditional feature extraction methods limit the ability on feature description.

With the development of deep learning and convolutional networks, different deep architectures have been proposed for FKV since 2016. Those architectures were distinguished as Siamese networks~\cite{li2016kinship} with a metric loss added at the end to guide the network training. Li~\cite{li2016kinship} proposed the Similarity Metric-based Convolutional
Neural Networks (SMCNN) method. Other metric-based loss functions, including contrastive loss and triplet loss were also commonly used~\cite{wu2022facial}. Moreover, the employment of extensive facial recognition models, pre-trained on comprehensive datasets, has notably enhanced the accuracy of kinship verification, offering a significant improvement in performance~\cite {shadrikov2020achieving}.

\begin{figure*}[!t]
\centering
\includegraphics[width=0.95\textwidth]{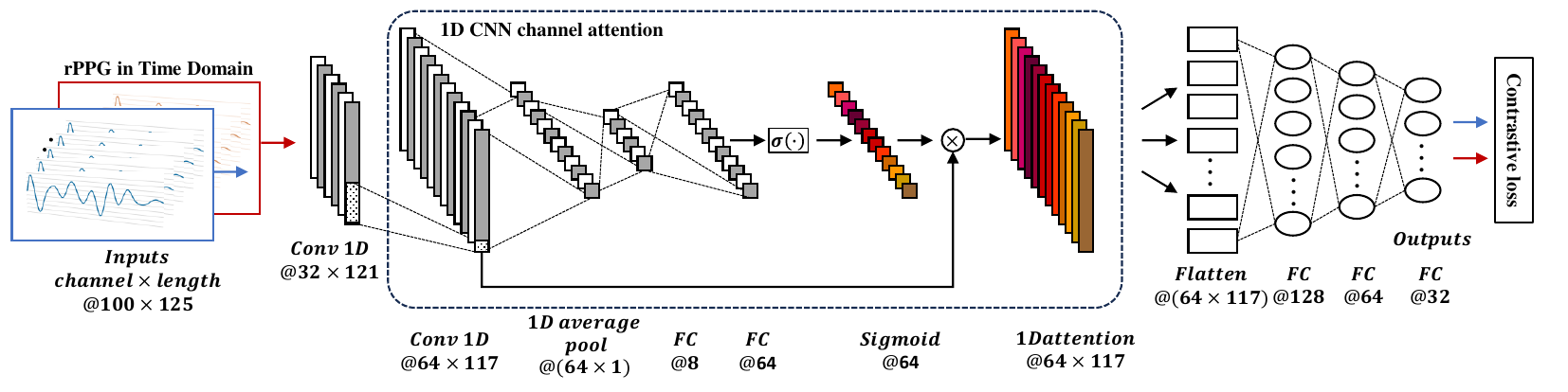}
\caption{Architecture of the proposed 1D-CNN channel attention model for kinship verification using facial rPPG. The model processes two rPPG inputs sequentially through two 1D convolution layers, a channel attention module, and three fully connected layers. ReLU serves as the activation function, and the Dropout technique is applied at the end to the first two fully connected layers. The network's training is guided by the contrastive loss, which is backpropagated for model updates.}
\label{fig:architecture}
\end{figure*}

Compared with traditional FKV from facial images or videos, the advantages of the use of rPPG can be concluded in three aspects. 
1) Privacy Preserving. Unlike conventional facial-based methods, rPPG eliminates the need for collecting facial images or videos. This not only prevents facial data storage but also mitigates emerging concerns on privacy regulations~\cite{perfail2024}. 
2) Security against Spoofing Attacks. Traditional biometric modalities, such as faces and voices, are susceptible to various threats~\cite{cai2021generative}, including spoofing and disguises. Heart biometrics, as rPPG signals in this paper, introduce a way of liveness detection, providing robust protection against data attacks.
3) Complexity. Compared with facial videos, rPPG contains much less and simpler 1D data depicting heart activity. The reduction from redundant information to a 1D signal significantly diminishes computational complexity, and the process could be done in real time with a normal computer~\cite{face2ppg}.

%\textbf{Heart Biometric} 

%%%%%%%%%%%%%%%%%%%%%%%%%%%%%%%%%%%%%%%%%%%%%%%%%%%

\vspace{0mm}
\section{rPPG Measurement from Facial Videos}
\vspace{0mm}
\label{sec:ppg}
Before we recognize the kinship, we first extract the remote photoplethysmography from the facial videos with an unsupervised learning manner. Given one video with a length of $L$ seconds, denoted as $v_i$, the measured vital sign from the $v_i$ is noted to be the $s_i$. A general workflow~\cite{boccignone2022pyvhr,face2ppg} for facial remote photoplethysmography (rPPG) measurement is illustrated in Figure \ref{fig:ppg}. 

It is noteworthy that current kinship datasets do not incorporate physiological signals measured via direct skin contact. Consequently, we integrate five non-machine learning-based rPPG methods from existing literature~\cite{nguyen2023non}: GREEN~\cite{verkruysse2008remote}, OMIT~\cite{face2ppg}, CHROM \cite{de2013robust}, LGI (Local Group Invariance) \cite{pilz2018local}, and POS (Plane-Orthogonal-to-Skin) \cite{wang2016algorithmic}. These conventional rPPG techniques capture variations in facial color caused by heartbeats and blood circulation.

\textbf{GREEN}~\cite{verkruysse2008remote} Among the three RGB color channels, the green channel has been shown to contain the most information to recover the rPPG. The GREEN method extracts the green color channel as the homonymous channel.

\textbf{OMIT}~\cite{face2ppg} recovers the rPPG by generating the orthogonal matrix with linearly uncorrelated components in the RGB signal, based on the matrix decomposition.

\textbf{CHROM} method \cite{de2013robust} derives rPPG signals by leveraging chrominance-related data. It creates two orthogonal chrominance signals from the original RGB data to reduce the impact of specular reflection. This reduction is achieved through a linear combination of color components based on a skin reflection model.

\textbf{LGI} \cite{pilz2018local} processes the initial signal and maps it into an invariant space to enhance the robustness of rPPG, especially when dealing with challenges like motion and varying lighting conditions.

\textbf{POS} \cite{wang2016algorithmic} introduces the concept of the Plane-Orthogonal-to-Skin. This technique projects the color space onto a plane that is orthogonal to skin color and then combines the results to generate the rPPG signal.

Due to the different color reflections of facial skin areas, in our implementation, we selected 100 rectangle facial Regions of Interest (ROIs) from the face to generate the remote photoplethysmography (rPPG). Those rPPG signals serve as the input to our network, consolidating data from various channels.

%The general pipeline for facial \hl{rPPG} measurement is shown in Fig.~\ref{fig:ppg}. In this paper, we study kinship correlations based on the rPPG. However, none of the existing kinship datasets includes the contact-measured physiological signals. So, we include three non-learning rPPG methods~\cite{nguyen2023non} in this paper: CHROM~\cite{de2013robust}, LGI (Local Group Invariance)~\cite{pilz2018local}, POS (Plane-Orthogonal-to-Skin)~\cite{wang2016algorithmic}. The traditional rPPG techniques measure the face color variance caused by the heart beat and blood pumping.

%\textbf{CHROM}~\cite{de2013robust} estimates the rPPG with chrominance-related information. It constructs two orthogonal chrominance signals from the original RGB signals to mitigate the impact of specular reflection. This is achieved through a linear combination of color components based on the skin reflection model.

%\textbf{LGI}~\cite{pilz2018local} projects the pre-processed signal into an invariance space, aiming at improving the rPPG robustness against the movement and lighting conditions.

%\textbf{POS}~\cite{wang2016algorithmic} introduces a Plane-Orthogonal-to-Skin that projects the color space into a plane orthogonal and fuses both to generate the rPPG.

%During the implementation, we choose 100 facial skin patches to generate the remote photoplethysmogram (PPG), which serves as the input to our network, incorporating various data channels.

\vspace{0mm}
%%%%%%%%%%%%%%%%%%%%%%%%%%%%%%%%%%%%%%%%%%%%%%%%%%%
\section{Method}
\label{sec:method}
\vspace{0mm}

In this section, we present a simple 1DCNN-Attention model tailored for kinship verification based on rPPGs. CNNs are designed to extract high-level convolutional features directly from raw data. While 2DCNNs are conventionally used for image processing, 1DCNNs are better suited for extracting spatial features from one-dimensional signals, like time series. Importantly, convolutional networks have demonstrated their proficiency in crafting effective feature representations from raw one-dimensional data, offering superior performance over deep multilayer perceptrons (MLP) without the need for intricate preprocessing~\cite{wang2017time}.
Incorporating an attention module allows for the calculation of channel-wise self-adaptive weights, thereby facilitating the learning of discriminative features. With 1DCNN serving as the feature extractor, the generated embeddings are subsequently projected through three fully connected layers. To capture and accentuate kinship traits, a contrastive loss is introduced at the network's end, guiding the training process. A visual representation of the entire architecture can be found in Fig.~\ref{fig:architecture}.

%In this section, we describe our proposed 1DCNN-Attention model for kinship verification from rPPGs. CNN is proposed to extract the high-level convolutional features from the raw data. While 2DCNN are typically used in image processing, 1DCNN are more suitable for extracting spatial features from one-dimensional signal, such as time-series. On the other hand, the convolutional networks have shown superior performance in crafting feature representations from raw one-dimensional data without complicated preprocessing~\cite{wang2017time} over deep multilayer perceptrons (MLP). Additionally, by applying the attention module, the channel-wised self-adaptive weights are calculated and discriminative features are learned. Since the 1DCNN is performed as the feature extractor, the embeddings are projected by three fully connected layers. Finally, to perceive the kinship traits, contrastive loss is added at the end of the network to guide the training. The whole architecture is shown in Fig.~\ref{fig:architecture}.

\vspace{0mm}
\subsection{1DCNN-Attention networks}
\vspace{0mm}
The training set for kinship verification is composed of multiple sample pairs. We denote them with $\mathcal{D}=\{(\mathbf{P}_i,\mathbf{C}_i,l_i),|i=1,2,\dots N\}$. $\mathbf{P}_i$ and $\mathbf{C}_i\in\mathbb{R}^{C\times W}$ represent the $i_{th}$ sample pair with rPPG generated by methods introduced in Section~\ref{sec:ppg}. $C$ and $W$ represent the number of input channels and the dimension of the rPPG signal. $l_i$ means the label for the $i_{th}$ sample pair, where $0$ represents that $\mathbf{P}_i$ and $\mathbf{C}_i$ are negative pairs without a kinship relation and $1$ denotes positive pairs with a kin relation. In this study, we intend to measure the kinship similarity of blood changes brought by the heartbeat. Therefore, for the input data configuration, we set the data sample as multiple channels to mitigate the color reflection effect of different facial ROIs.

For a Siamese input, the rPPG data pair $(\mathbf{P}_i,\mathbf{C}_i)$ is forwarded to the network with the same strategy. Let's take $\mathbf{P}_i$ as an example. We designed two 1D convolution blocks. $\mathbf{P}_i$ first passed two convolutional operations with 1-D kernels with the size of \{5, 5\}, stride \{1, 1\} without padding, and output channels are \{32, 64\}, respectively. The convolutional computation block can be represented as:
\begin{equation}
    \left\{\begin{matrix}\mathbf{y}  = \mathbf{W} \otimes \mathbf{x} +b
 \\\mathbf{h}  = ReLu(\mathbf{y})

\end{matrix}\right.
\end{equation}
where $\otimes$ is the convolutional operation, $\mathbf{x}$ is the input to the convolutional block, and $\mathbf{h}$ is the obtained feature embedding by applying the ReLU activation function.

\textbf{Channel Attention module} To locate the discriminative kinship features, we further utilize a 1D self-attention module. Let $\hat{\mathbf{P}_i}\in\mathbb{R}^{\hat{C}\times \hat{W}}$ be the feature collected at the second convolutional block. $\hat{C}$ and $\hat{W}$ represent the number of channels and spatial dimensions. To squeeze the global information into channel-wise data distribution~\cite{hu2018squeeze}, a 1D global average pooling is adopted. Following that, there are two FC layers with a reduction ratio of $r$. $r$ is set to 8. The channel-wise weight vector and output of the attention module can be calculated by:
\begin{equation}
    \mathbf{w}_p = \sigma\left(FCs(g(\hat{\mathbf{{P}}_i}))\right), \mathbf{z} = \hat{\mathbf{{P}}_i}\mathbf{w}_p
\end{equation}
$\sigma(\cdot)$ is a Sigmoid function, $g(\hat{\mathbf{{P}}_{i}})=1 /\hat{W} \sum_{i=1}^{\hat{W}} \hat{\mathbf{{P}}_{ij}}$ is the 1D global average pooling. $FCs(\cdot)$ denotes two FC layers. $\mathbf{z}$ has the same size as $\hat{\mathbf{P}_i}$. The final output can be calculated by:
\begin{equation}
    \left\{\begin{matrix} \mathbf{f} =FLATTEN(\mathbf{z} )
 \\\mathbf{f} _1 = DropOut(ReLu(FC1(\mathbf{f} )))
 \\\mathbf{f} _2 = DropOut(ReLu(FC2(\mathbf{f} _1)))
 \\\mathbf{f}_p^i  = FC3(\mathbf{f} _2 )

\end{matrix}\right.
\end{equation}
$FLATTEN(\cdot)$ transforms the feature map $\mathbf{z}$ into a one-dimensional vector. The dropout rate is set at 0.1. After passing three FC layers, we can collect the feature output as $\mathbf{f}_p^i$. Similarly, for the other network input $\mathbf{C}_i$, the feature represented by it can be denoted as $\mathbf{f}_c^i$.

\vspace{0mm}
\subsection{Loss function}
\vspace{0mm}
To train the 1DCNN network, we implement the contrastive loss to optimize the model. The network outputs the pair-wise embedding as $\{(\mathbf{f}_p^i,\mathbf{f}_c^i,l_i),|i=1,2,\dots N\}$, where $\mathbf{f}_p^i$ and $\mathbf{f}_c^i$ are collected at the last FC layer of the network and $l_i$ denotes ground truth with 1 denoting kinship ad 0 the otherwise. To force the model learning the robust rPPG feature representations for kinship verification, we minimize the contrastive loss below:
\begin{equation}\begin{matrix}
\mathcal{L_{rPPG}}=\frac{1}{N}\sum_{i=1}^{N} (l_i d_i^{2} + (1-l_i)\max(\tau -d_i,0)^2)
 \\s.t. \quad d_i=\left \| \mathbf{f}_p^i-\mathbf{f}_c^i \right \|^{2}
\end{matrix}
\end{equation}
$\tau$ is the hyperparameter, and $d_i$ denotes the Euclidean distance between a sample pair.

\vspace{0mm}
\section{Experiments}
\label{sec:experiment}
\vspace{0mm}

\begin{table*}[!t]
\centering
\normalsize
\caption{Evaluation of our method on the UvANEMO Smile Database for seven kin relations using AUC (\%). Different rPPG methods were employed. Bold numbers indicate the best results for each relation.}
\vspace{-1mm}
\begin{tabular}{cc|cccc|ccc|c}
\hline
\multicolumn{2}{c|}{\multirow{2}{*}{Kin relations}} & \multicolumn{4}{c|}{Parent-Child} & \multicolumn{3}{c|}{Siblings} & \multirow{2}{*}{Mean$\pm$Std.} \\ \cline{3-6} \cline{7-9}
\multicolumn{2}{c|}{}& F-S    & F-D    & M-S    & M-D& B-B& S-B& S-S& \\ \hline \hline
\multirow{3}{*}{{1Dattention}}         & LGI~\cite{pilz2018local} & {59.91}  & 59.79  & 53.11  & 57.43  & 56.14    & 64.25    & 75.55   & 60.88$\pm$6.80  \\ \cline{2-10}
& OMIT~\cite{face2ppg} & 61.55 & 63.00 & 58.31 & 50.68 & 69.17 & 70.12 & 79.73 & 64.65 $\pm$8.66 \\ \cline{2-10} 
& GREEN~\cite{verkruysse2008remote} & 56.92 & 61.00 & 67.47 & 56.32 & 75.19 & 77.12 & 81.48 & 67.93$\pm$9.45 \\ \cline{2-10} 
& CHROM~\cite{de2013robust}        & 60.36  & {60.14}  & 62.44  & 52.53  & \textbf{82.46}    & \textbf{82.28}    & \textbf{81.53}   & 68.82$\pm$11.84           \\  \cline{2-10}
& POS~\cite{wang2016algorithmic} & \textbf{70.43}  & \textbf{68.61}  & \textbf{68.94}  & \textbf{70.42}  & 61.65    & {72.43}    & {72.47}   & \textbf{69.28$\pm$3.41}\\
\hline
\end{tabular}
\label{table:ppg}
\end{table*}

\begin{table*}[!t]
\centering
\normalsize
\caption{Ablation study of the proposed architecture on multichannel inputs and 1D channel attention using AUC (\%). CA means the network channel attention. SingleChannel indicates that the input is one rPPG signal per facial video. The colored cells show the performance comparison of the proposed and ablated methods.}
\vspace{-1mm}
\begin{tabular}{c|cccc|ccc|c}
\hline
\multirow{2}{*}{Kin relations} & \multicolumn{4}{c|}{Parent-Child} & \multicolumn{3}{c|}{Siblings} & \multirow{2}{*}{Mean$\pm$Std.} \\
\cline{2-5} \cline{6-8}
& F-S  & F-D & M-S & M-D & B-B& S-B& S-S& \\
\hline
\hline
SingleChannel & 66.35 &	66.43 &	66.14	&71.18	&53.00	&66.70	&71.43&	65.89 $\pm$6.14\\
\hline
w/o. CA & 69.71  & 64.43  & 66.93  & 71.22 & 60.40    & 72.29   & 68.35   & 67.62$\pm$4.13 \\
\hline
Proposed & 70.43  & 68.61  & 68.94  & 70.42 & 61.65    & 72.43   & 72.47   & 69.28$\pm$3.69 \\
\hline
\rowcolor{lavender}
\cellcolor{white} (Multi-channel) over (Single-channel)
& $\uparrow$\textbf{4.08}   & $\uparrow$\textbf{2.18}   & $\uparrow$\textbf{2.80}   & $\downarrow${0.76} & $\uparrow$\textbf{8.65} & $\uparrow$\textbf{5.73}    & $\uparrow$\textbf{1.04}   & $\uparrow$\textbf{3.39}\\
\hline
\rowcolor{lavender}
\cellcolor{white} (w/ CA) over (w/o. CA) & $\uparrow$\textbf{0.72}   & $\uparrow$\textbf{4.18}   & $\uparrow$\textbf{2.01}   & $\downarrow${0.80} & $\uparrow$\textbf{1.25} & $\uparrow$\textbf{0.14}    & $\uparrow$\textbf{4.12 }   & $\uparrow$\textbf{1.66}\\
\hline
\end{tabular}
\label{table:ablation}
\end{table*}

\subsection{Implementation Details}
\vspace{0mm}
\begin{figure}[!t]
\centering
\includegraphics[width=0.47\textwidth]{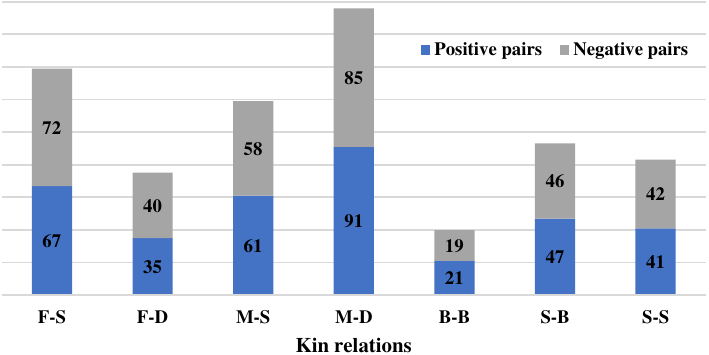}
\centering
\caption{Data distribution within each kin relation in the subset of UvA-NEMO Smile Database~\cite{dibekliouglu2012you}.}
\label{fig:Smile}
\vspace{0mm}
\end{figure}
\textbf{Dataset and evaluation metric.} The UvANEMO Smile Database is widely employed for kinship verification studies~\cite{dibekliouglu2012you}. It comprises front-facing videos of spontaneous/posed smiles captured in a constrained indoor environment. These videos have a resolution of 1920 × 1080 pixels and a frame rate of 50 FPS. The database contains 1240 smile videos from 400 subjects spanning seven kin relations~\footnote{Parent-Child kin relations: Father-Son (F-S), Father-Daughter (F-D), Mother-Son (M-S), and Mother-Daughter (M-D); Sibling kin relations: Brother-Brother (B-B), Sister-Sister (S-S), and Sister-Brother (S-B)}. Videos range in duration from 1.1 to 14.3 seconds. To ensure a sufficient time window for generating rPPG signals, we focus on videos lasting at least 2.5 seconds. From this subset, we selected both negative and positive pairs meeting the length criteria. The distribution of the seven kin relations within the data is depicted in Fig.~\ref{fig:Smile}.
In our experiments, we evaluate all seven relationship types. The primary metric is the area under the receiver operating characteristic (AUC), with the ROC curve visualizing model performance.
%\textbf{Dataset and evaluation metric.} The UvANEMO Smile Database has been widely employed in the kinship verification study~\cite{dibekliouglu2012you}. It was collected under a constrained indoor environment consisting of front-facing videos of spontaneous/posed smiles. The videos are with a resolution of 1920 $\times$ 1080 pixels. The frame rate is 50 FPM (frames per second). There are a total of 1240 smile videos from 400 subjects, and seven kin relations are included~\footnote{Parent-Child kin relations: Father-Son (F-S), Father-Daughter (F-D), Mother-Son (M-S) and Mother-Daughter (M-D); Sibling kin relations: Brother-Brother (B-B), Sister-Sister (S-S) and Sister-Brother (S-B)}. The duration of the videos is between 1.1 seconds to 14.34 seconds. To ensure enough time window size for generating the rPPG signals, we selected a subset with a video length of no less than 2.5 seconds. Based on that, we also selected the negative pairs and positive pairs that satisfied the length requirement. The data distribution of seven kin relations is shown in Fig.~\ref{fig:Smile}.
%We evaluate 7 relationship types in our experiments. The area under the receiver operating characteristic (AUC) is the primary metric, and the ROC curve (receiver operating characteristic curve) is used to illustrate the model performance.

\textbf{Training details.} To validate the effectiveness of the proposed method, we apply a leave-one-subject-out (LOSO) cross-validation scheme~\cite{dibeklioglu2017visual}. In this scheme, one subject pair is reserved for testing, while the remaining data for each kin relation is used for training. During the training of the proposed 1D CNN, we allocate a portion of the training set as a validation set to monitor the training process, using early stopping as needed. The network parameters are optimized using the Adam optimizer with a learning rate of 1e-3 and a mini-batch size of 30. The model is trained using the PyTorch framework on an Nvidia Tesla V100 GPU (16 GB). The rPPG methods are implemented using Face2PPG, a modified version of the pyVHR toolbox 2.0~\cite{boccignone2022pyvhr,face2ppg}\footnote{\url{https://github.com/phuselab/pyVHR}}. We set the parameter $\tau$ to 1. %The source code can be found from: \url{github.com/}.
%[MIGUEL] If you do not have the code available you can add a placeholder herei if you want

\vspace{0mm}
\subsection{Results and analysis}
\vspace{0mm}
Table~\ref{table:ppg} summarizes the results of the proposed 1DCNN model with the inputs of rPPG signals generated by the LGI~\cite{pilz2018local}, OMIT~\cite{face2ppg}, GREEN~\cite{verkruysse2008remote}, CHROM~\cite{de2013robust}, POS~\cite{wang2016algorithmic} methods, respectively. On average, the model with the inputs of rPPG by POS~\cite{wang2016algorithmic} method yields the best recognition performance, which shows the most discriminative ability. For the problem of FKV in relation of Parent-Child, POS~\cite{wang2016algorithmic} rPPG recover method shows the best performance. While for the case of siblings, the CHROM~\cite{de2013robust} method outperforms others. In this work, we primarily study the rPPG for the FKV problem based on traditional rPPG methods without label training. Nevertheless, these findings can demonstrate the discrimination within rPPG signals for kinship recognition, answering the question posed in this paper.

Additionally, we visually compare the performances of different rPPG methods by displaying their respective ROC curves in Fig.~\ref{fig:ppgroc}. This visual representation reinforces the notion that rPPG signals from the POS method offer the highest level of discrimination.% Fig.~\ref{fig:loss} shows the loss changes over training epoches.

%\begin{figure}[htb]
%\centering
%\includegraphics[width=0.47\textwidth]{loss.eps}
%\centering
%\vspace{0mm}
%\caption{Training loss over training epoches.}
%\label{fig:loss}
%\end{figure}
\vspace{0mm}
\subsection{Ablation Study}
\begin{figure}[!t]
\centering
\begin{subfigure}[b]{0.23\textwidth}
         \centering
         \includegraphics[width=\textwidth]{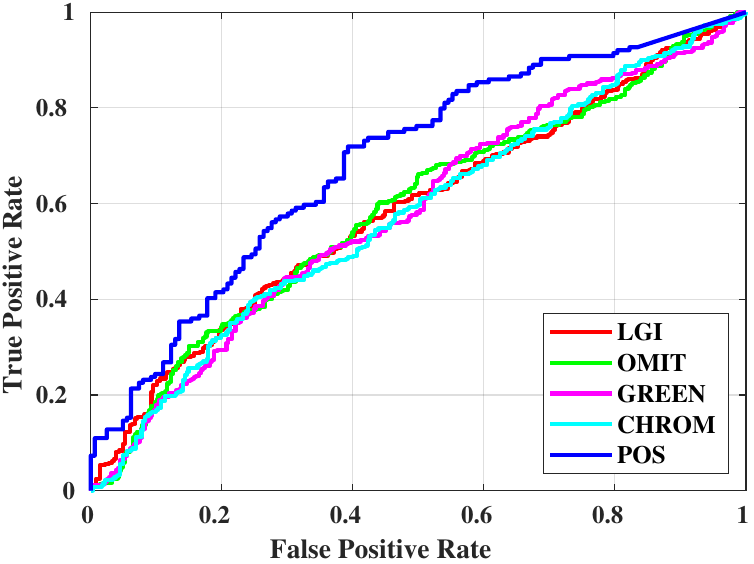}
         \caption{rPPG methods}
         \label{fig:ppgroc}
     \end{subfigure}
\begin{subfigure}[b]{0.23\textwidth}
         \centering
         \includegraphics[width=\textwidth]{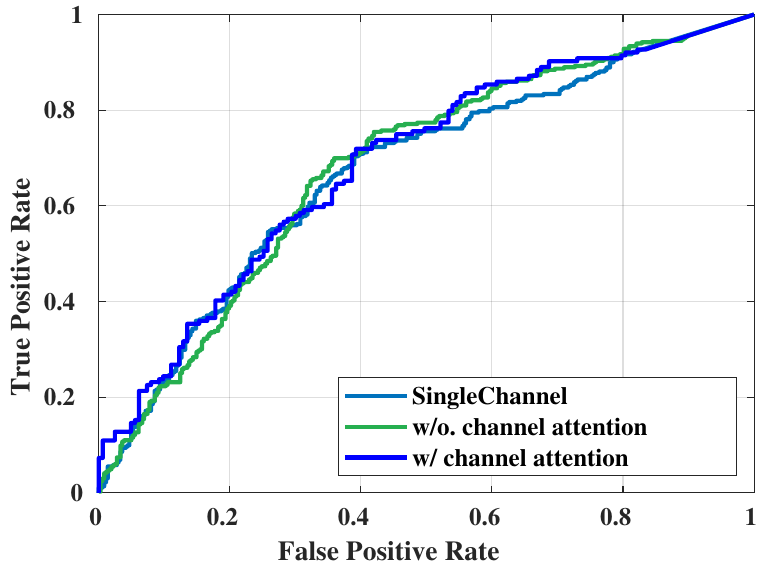}
         \caption{Ablation study}
         \label{fig:ablroc}
     \end{subfigure}
\vspace{0mm}
\caption{(a) The ROC curves comparison between different inputs by rPPG methods.  (b) The ROC curves comparison for the ablation study. Zoom in for better visualization.}
\vspace{0mm}
\label{fig:train}
\end{figure}

\vspace{0mm}
We conducted an ablation study based on our proposed model using rPPG inputs generated by the POS method~\cite{wang2016algorithmic}. The results of this study are detailed in Table~\ref{table:ablation}, and the corresponding ROC curves are illustrated in Fig.~\ref{fig:ablroc}. The final two rows of the table provide a performance comparison for the two ablation scenarios.

%Based on the proposed model with the inputs of rPPG generated by POS~\cite{wang2016algorithmic} method, we implemented the ablation study. The experimental results are summarized in Table~\ref{table:ablation} with the corresponding ROC curves illustrated in Fig.~\ref{fig:ablroc}. The last two lines show the performance comparison on two ablations.

\textbf{Comparison with Single Channel Input.}
Instead of utilizing a 100-channel rPPG input, we also tested a single-channel rPPG derived from one facial video using the holistic method from Face2PPG/pyVHR~\cite{boccignone2022pyvhr,face2ppg}. All other model components were kept consistent. As seen from the penultimate row in Table~\ref{table:ablation}, multi-channel rPPG inputs enhance the model's robustness, yielding an average performance improvement of 3.39\%.

%Instead of using the 100-channel rPPG as the input, we further generated the one channel rPPG from one facial video by using the holistic method in pyVHR~\cite{boccignone2022pyvhr,Boccignone2020} as the input to the model while keeping other components unchanged. From the second last line in table~\ref{table:ablation}, we can find that the multi-channel input of rPPG helps to improve the robustness of the model and could bring an average performance improvement with 3.39\%.

\textbf{Effectiveness of Channel Attention.}
We evaluated the impact of the channel attention module in our model, as presented in Table~\ref{table:ablation}. ``w/o. CA" denotes a version of the network without the 1D CNN channel attention module. As indicated in the last row of Table~\ref{table:ablation}, incorporating the channel attention module results in an average 1.66\% AUC boost, underscoring its significance in the model.

%We further evaluate the effectiveness of the channel attention module in Table~\ref{table:ablation}. The w/o. CA denotes that the network removes the 1D CNN channel attention. The last line of Table~\ref{table:ablation} compares the performance. The channel attention module brings about 5.5\% AUC gains, which demonstrates the necessity of it.
\vspace{0mm}
\section{Conclusion}
\vspace{0mm}
In this paper, we focus on addressing the kinship verification problem through the rPPG signals extracted from facial videos. Firstly, five traditional rPPG methods were implemented to recover the rPPG. Then, we utilize a 1DCNN network with a self-attention module to learn the kinship traits from the obtained rPPG. It leverages multiple rPPG as multi-channel inputs and incorporates the 1D attention module to extract discriminative kin features. Our ablation experiments demonstrate that the rPPG POS method yields the highest kinship discriminative ability, highlighting the effectiveness of the raised model components. The shortcoming of this paper lies in that the robustness of our system with noisy kinship data remains unexplored, subject to the robustness of rPPG methods such as supervised-learning methods. As part of our future work, we intend to delve into this aspect, further enhancing the applicability of our approach.

% \vfill\pagebreak
% \clearpage

\bibliographystyle{IEEE.bst}
\bibliography{ref.bib}

\end{document}